# Sentiment Analysis Dataset in Moroccan Dialect: Bridging the Gap Between Arabic and Latin Scripted dialect.

JBEL MOUAD*

University Sultan Moulay Slimane, Beni-Mellal, Morocco, mouad.jbel@gmail.com

JABRANE MOURAD

University Sultan Moulay Slimane, Beni-Mellal, Morocco, mourad.jabrane@usms.ac.ma

HAFIDI Imad

University Sultan Moulay Slimane, Beni-Mellal, i.hafidi@usms.ma

METRANE Abdulmutallib

University Sultan Moulay Slimane, Beni-Mellal, ab.metrane@gmail.com

## Abstract

Sentiment analysis, the automated process of determining emotions or opinions expressed in text, has seen extensive exploration in the field of natural language processing. However, one aspect that has remained underrepresented is the sentiment analysis of the Moroccan dialect, which boasts a unique linguistic landscape and the coexistence of multiple scripts. Previous works in sentiment analysis primarily targeted dialects employing Arabic script. While these efforts provided valuable insights, they may not fully capture the complexity of Moroccan web content, which features a blend of Arabic and Latin script. As a result, our study emphasizes the importance of extending sentiment analysis to encompass the entire spectrum of Moroccan linguistic diversity.

Central to our research is the creation of the largest public dataset for Moroccan dialect sentiment analysis that incorporates not only Moroccan dialect written in Arabic script but also in Latin letters. By assembling a diverse range of textual data, we were able to construct a dataset with a range of 20.000 manually labeled text in Moroccan dialect and also publicly available lists of stop words in Moroccan dialect.

To dive into sentiment analysis, we conducted a comparative study on multiple Machine learning models to assess their compatibility with our dataset. Experiments were performed using both raw and preprocessed data to show the importance of the preprocessing step. We were able to achieve 92% accuracy in our model and to further prove its liability we tested our model on smaller publicly available datasets of Moroccan dialect and the results were favorable.

---

* Corresponding author.  *E-mail address:* mouad.jbel@gmail.com

# 1.Introduction

The field of sentiment analysis (SA), a subdomain of natural language processing (NLP) [1], has witnessed remarkable progress in recent years, enabling us to understand and analyze human sentiments expressed through text data. SA holds great promise for various applications, including social media monitoring, market research, customer feedback analysis, and political sentiment tracking. While substantial progress has been made in analyzing sentiment in major world languages [2,3,4], the exploration of SA in dialectal languages remains an underexplored and challenging terrain. In this research, we embark on a journey to bridge this gap by delving into SA in the Moroccan dialect (MD). This study is significant for several reasons, notably because it explores sentiments expressed not only in Arabic script but also in Latin letters, which is a common occurrence on the internet.

Moroccan dialect, a regional variant of the Arabic language, serves as a vital medium of communication in Morocco. It plays a pivotal role in reflecting the sentiments, attitudes, and opinions of Moroccan individuals across a wide range of contexts. Analyzing sentiment in MD can offer invaluable insights into various aspects of Moroccan society, from political discourse to customer preferences in the local market. However, the unique linguistic characteristics of MD pose challenges for sentiment analysis, making it an intriguing and understudied subject within the field of NLP.

Sentiment analysis in MD is particularly challenging due to its distinctive linguistic features [5,6], including pronunciation variations, colloquial expressions, and borrowed words from French, Spanish, and Berber languages. Furthermore, the use of multiple scripts, primarily Arabic and Latin letters, adds complexity to SA. While Arabic script is traditionally used for written MD, Latin letters have become increasingly popular, especially in online communication. This transition from Arabic script to Latin letters, driven by the digital age and the prevalence of social media, has created a unique challenge for researchers in the field of SA.

Previous research in SA has primarily focused on dialects written or coded using Arabic letters only [11,17]. While this approach has yielded valuable results, it becomes apparent that the landscape of the internet, where much of the data for SA is sourced, includes a diverse range of linguistic representations. As we aim to analyze sentiments expressed on the web, focusing solely on Arabic script is not entirely relevant to the data we encounter. This limitation diminishes the ability of existing SA models to fully capture and understand the nuances present in MD data. Existing SA models designed for Arabic script may struggle when applied to text written in Latin letters, as they may not adequately account for the phonetic and orthographic variations introduced by this script shift. Therefore, there is a pressing need to adapt and expand SA techniques to encompass the full spectrum of linguistic diversity present in MD texts found on the web.

To address these challenges, this research aims to create the first and largest public dialect dataset for Moroccan SA that aims at both scripting types. This dataset will be a crucial resource for advancing the understanding and analysis of sentiment in the MD across multiple scripts. It includes a diverse collection of texts, encompassing both Arabic script and Latin script, thus reflecting the evolving linguistic landscape of online content in Morocco. Our dataset is carefully created, ensuring its quality and representativeness, and to push even further we managed to create stop words lists for both type of scripts that will also be publicly available, and will serve as a valuable resource for researchers and practitioners interested in SA in dialectal languages.

A critical aspect of this research is the identification and selection of machine learning (ML) models that are best suited to our newly curated dataset. The goal is to develop models that can effectively capture the nuances and subtleties of the MD, providing accurate SA results. This process will entail a thorough exploration of various ML models, including deep learning (DL) approaches such as Long short-term memory (LSTM) and convolutional neural networks (CNN), as well as traditional ML algorithms like support vector machines (SVM), K-nearest neighbors (KNN), and Naïve Bays (NB).

Model selection will be guided by the principles of accuracy, robustness, and adaptability. Our approach will involve the training and evaluation of these models on our comprehensive dialect dataset, with a particular emphasis on their performance in handling texts written in both Arabic and Latin scripts. This iterative process will help us identify the ML model that excels in SA of the MD, setting a benchmark for future research in this domain.

In addition to model selection and training, this research will evaluate the generalizability and performance of the chosen SA model on other existing public Moroccan datasets, and their evaluation will serve as a critical step in assessing the versatility and applicability of our model. By conducting these cross-dataset assessments, we aim to determine the models to treat both types of scripted MD since all other available datasets are only written in Arabic script. This comprehensive evaluation will provide insights into the strengths and limitations of our model and dataset, shedding light on its real-world utility beyond the MD. It will contribute to the ongoing efforts to enhance SA techniques for dialectal languages and facilitate knowledge sharing within the research community.

In summary, this research represents a significant step towards advancing the field of SA by addressing the unique challenges posed by the MD, including the use of multiple scripts. Our contributions include the creation of a comprehensive dialect dataset, the identification of the most suitable ML model for SA in MD, and the evaluation of its performance on other publicly available datasets. Through these efforts, we aspire to provide valuable insights into the sentiment dynamics of Moroccan online discourse, with broader implications for the analysis of dialectal languages and cross-script SA on the web. This research holds the potential to inform applications across various domains, from understanding public opinion to enhancing customer experience analysis in the Moroccan context.

we have structured this paper as follows: the second section of the paper presents similar works and existing datasets in dialects, Modern Standard Arabic (MSA) and MD. The third section discusses MD challenges in SA. The fourth section is devoted to presenting the steps we planned to create our dataset. In the fifth segment, we discuss the ML experiments and models we use. And the sixth section is dedicated to discussing the results of our models.

## 2. Related works

Since Arabic is the fourth most widely spoken language and a language with a fast-growing user base on the Internet, SA in Arabic is regarded as an important research activity in the field of SA. However, because Arabic is a morphologically complex language, SA for Arabic is thought to be challenging. Many studies of SA and Arabic datasets have conducted on Arabic written text in the recent years [7,8,9], although most of them focused on MSA, but there is also a few who done research on Arabic dialects [10, 11,12].

In the following sections, we will introduce various datasets and corpora generated by different research communities in the field of SA. These resources can be categorized into three distinct groups: those related to MSA, datasets originating from Arabic dialects other than Moroccan, and finally, all available resources designed for MDs to our best knowledge.

*MSA datasets:*

Many MSA datasets have been created, the most famous of which are AWATIF [29] and OCA [30]. The first is a multi-genre corpus extracted from three different sources, including the Penn Arabic Treebank (PATB), as well as Wikipedia web forum and discussion pages, with a total of 5,382 sentences. The researchers removed noise and text written in dialects to retain only the MSA, and applied a manual annotation process to compare labels derived from native speakers with labels obtained from the system Crowdsourced from Amazon Mechanical Turk. Annotators used principles that were both linguistic and nuanced.

Regarding OSA, reviews are manually extracted directly from various movie websites. There are a total of 500 OCA reviews, including 250 positive reviews and 250 negative reviews. Manual processing for text cleaning was performed while Rapidminer software was used to evaluate both Khoja and light searching.

Another known dataset is ArSAS[13] an Arabic tweet corpus that has been annotated for SA tasks. 21k Arabic tweets covering a wide range of topics were compiled, prepared, and annotated in a sizable collection. Twenty diverse themes from around the world that are anticipated to elicit spirited discussion among Twitter users were used to extract and collect the tweets in the corpus. The tweet collection did not rely on emotions or sentiment keywords, especially for the task of SA, to avoid data bias to a given lexicon.

HARD (Hotel Arabic-Reviews Dataset) [14] is too a MSA dataset of Arabic book reviews for arbitrary SA and ML applications. 490587 hotel reviews were compiled by HARD from Booking.com. The Arabic review text, the reviewer's star rating out of ten, and other details about the hotel and the reviewer are all included in each record. They used six popular classifiers to examine the datasets. The classifiers were put through their paces in terms of polarity and rating classification.

*Dialects datasets:*

SANA, a massive multilingual lexicon for SA built on both MSA and informal Arabic, was created by Abdul-Mageed and Diab [9] (Egyptian and Levantine). The MSA and several Arabic dialects are covered by the 224,564 entries in the SANA. The authors manually labeled two-word lists from the Yahoo Maktoob and Penn Arabic Treebank. they used Google's translation API to generate lists of three existing English lexica: SentiWordNet, YouTube Lexicon, and

GeneralInquirer. To increase the lexicon's coverage, they employed a statistical technique based on PMI to extract additional polarized tokens from the Twitter and conversation datasets. Despite the resource's magnitude, many of the items are neither lemmatized or dia-critized, which restricts the

[15] Rahab et al. put forth a technique for classifying Arabic comments taken from websites belonging to Algerian newspapers as positive or negative. For this study, they created an Arabic corpus known as SIAAC (Sentiment polarity Identification on Arabic Algerian newspaper Comments) of 150 entries. They tested Support Vector Machines (SVM) and Naive Bayes, two popular supervised learning classifiers (NB). To compare and assess the outcomes of their trials, they used a variety of factors and metrics (recall, precision and F-measure). The best results in terms of precision were generated by SVM and NB. It has been demonstrated that using a bi-gramme improves the accuracy of the two models. Additionally, SIAAC displayed competitive results when compared to OCA.

[31] presented a Lexicon-Based SA Approach for Saudi dialect in twitter. A manually created lexicon with a list of words and with their assigned polarities and some common phrases were proposed and tested for negative and positive classification. Compared to the bigger automatically created dictionary AraSenTi, SauDiSenti's SA for tweets in the Saudi dialect demonstrated impressive results despite its small size. The performance of SauDiSenti can be enhanced by including dictionaries and introducing new themes.

*Moroccan dialect Datasets:*

A framework for Moroccan SA was created by Oussous, A. et al[11] where they implemented many techniques for SA. The authors build a dataset that consisted of 2000 Moroccan text annotated for positive and negative under the name of MSAC. They were able to achieve promising results considering the small dataset size where their model scored an accuracy of 99%. However, the study only focused on text written in Arabic letters and didn't include a test on dialect scripted in Latin letters.

[16] This paper enhances the Arabic resources landscape by introducing a substantial Moroccan dataset sourced from Twitter, meticulously annotated by native speakers. This dataset, extracted from Twitter, includes supplementary categories, including objective text and text conveying sarcasm. To assess its performance, the dataset underwent testing with ML models, yielding encouraging outcomes.

[17] Is another work that focus on analyzing Facebook comments that are expressed in modern standard or in Moroccan dialectal Arabic. They managed to create a dataset under the name of MAC that contains 3542 MD text. To test the dataset they created two models, one that considers all the Arabic text as homogenous and the second one that require a dialectical detection beforehand SA.

Another work that focus on MD is done by Errami et al[18], where they managed to create a dataset dedicated to SA from Moroccan tweets, it consists of 3200 tweets scripted with Arabic letters and labeled into two classes positive and negative. To test their dataset a model was created using LSTM, the results were promising.

## 3.Challenges of Sentiment analysis in Arabic dialects

This section discusses some of the major challenges in Arabic dialects Sentiment Analysis.

**Morphological analysis:** Decomposing words into morphemes and assigning each morpheme morphological information, such as stem, root, POS (Part of Speech), and affix, is the basic goal of morphological analysis. It is significantly more challenging to complete these tasks in Arabic because of the language's intricate morphology. This complexity necessitates the creation of appropriate systems capable of dealing with tokenization, spell checking, and stemming.

**Scarcity of datasets and corpora:** Compared to English, web content in Arabic is rare and there are few datasets available to apply SA to the Arabic language. Therefore, because the accuracy of SA depends on the amount of data, it is difficult to compare performance between Arabic and other languages.

**Complexity of dialects**: Arabic speakers typically use colloquial Arabic rather than MSA for communication (Moroccan dialect in our case). There are no language academies or standard orthographies for MD. As a result, processing Arabic dialects with methods and resources made for MSA yields noticeably subpar outcomes. Recently, researchers started creating parsers for certain dialects [19,20]. These assessments are only made for specific dialects and still have poor accuracy. The efficiency of information retrieval will increase once this gap in Arabic processing is closed, particularly for data from social media.

**Dialect scripted in Latin letters:** Latin-based Arabic writing system is widely used to write MSA and Arabic dialects on the web [21], this can cost a lot when it comes to accuracy and precision of detecting the polarity. Our work focuses on bridging this gap by offering resources of data scripted in Latin letters too. table 1 show an example of this writing technique.

Table 1. Examples of different encodings.

| Encoding class | Text |
|---|---|
| English | *Well done brother, can you give me the name of the application?* |
| MSA | أحسنت اخي هل يمكن تعطيني اسم التطبيق ؟ |
| Moroccan dialect in Arabic letters | مزيان خويا واش تقدر تعطيني اسم تطبيق؟ |
| Moroccan dialect in Latin letters | *Mzyan khouya, t9dr t3tini smiya dyal tatbi9 ?* |

## 4.Moroccan YouTube Corpus (MYC)

*Dataset creation steps:*

If this area of research is to advance, tools and resources are required. The procedure for gathering the MD raw text for our dataset is described in this section. In contrast to MSA, which predominates in written resources like news media, education, science, and books, MD is only utilized in informal circumstances like dialogues in TV series and movies. The past few years, written MD has begun to develop in social media platforms (Facebook, Twitter, blogs, etc.). MD delivers socially motivated commentary on various areas and issues, from personal narratives to traditional folk literature, even though it is employed in informal settings (stories, songs, etc.). It was incredibly challenging to locate and gather resources for MD. The lack of a standardized orthography, the existence of various subdialects, and the widespread use of different writing scripts (Arabic vs. Latin) make MD resources susceptible to significant noise and inconsistency, which makes it difficult for techniques using query matching to identify dialectal text in the particular dialect of interest.

In our dataset[1] we focus on variety too rather than on mere size. That's why we chose YouTube videos with different topics to extract their comments. Our approach to collecting MD comments is described as follows:

- Creating a script using YouTube API to extract comments from YouTube channels.
- Manually choosing 50 Moroccan famous YouTube channel of different topics as targets to ensure the variety of resources, then select and determine suitable MD content. The selection and reviewing of resources were performed by native MD speakers.
- Moroccans tends to use French and English in social media platforms, therefor we performed a cleaning phase where we delete content that contains only foreign languages, but we made sure that our dataset include the Latin version of MD.
- Manual annotation of comments into two classes, positive comments and negative comments, was performed by five native MD annotators; two of them were experts who worked in social media project management.
- The voting technique was chosen to determine the comments polarity, we agreed that the numbers of vote required to decide the comment's class is four votes out of five, this way we ensure that the error margin is at its lowest.

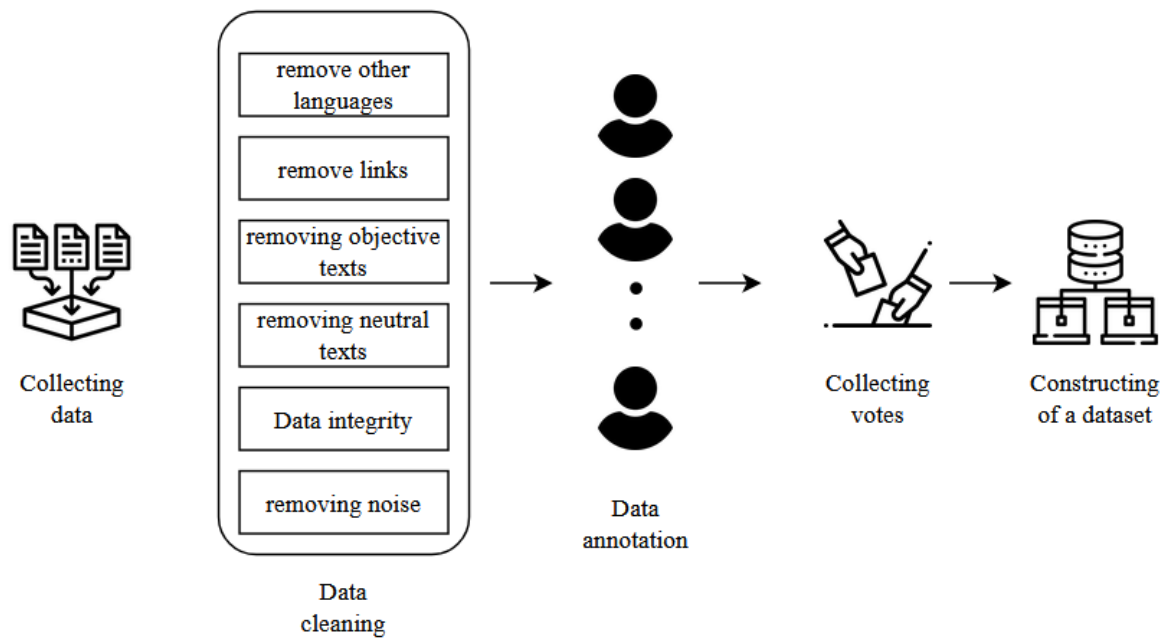

Fig. 1. Dataset creation steps

Figure 1 depicts a schema of the steps described previously. We collected comments posted between 2018 and 2022 obtained from different topics and locations in Morocco. Then, those comments are used to perform polarity classification.

---

[1] https://github.com/MouadJb/MYC

**Stop words lists:**

In addition to the dataset, we carefully created lists of the most commonly used stop words in the web to further improve the step of preprocessing in our model, since many previous studies demonstrate that the removal of stop words can have huge impact on the task of SA [22]. The lists that contain more than 350 words will also be publicly available as a contribution for this field of research. Table 2 give an insight of the two lists.

Table 2. Examples of Moroccan dialect stop words.

| Moroccan dialect stop words lists | |
|---|---|
| Stop words scripted in Arabic letters | Stop words scripted in Latin letters |
| ['ila','3a','ta','idan','9al','chwiya','ktar','bzaf','chi',…] | ['أقل','حيت','نتا','معانا','على', 'أكثر','ألا',…] |

*Dataset contents:*
the dataset was built of twenty thousand annotated and validated comments for sentiments, and it has ten thousand positive comment and ten thousand negative comments. Each of the ten thousand comments were balanced between comments written in Arabic letters and comments written in Latin letters. Figure 2 describes the content of the dataset more clearly and table 3 give an example of each type of comments.

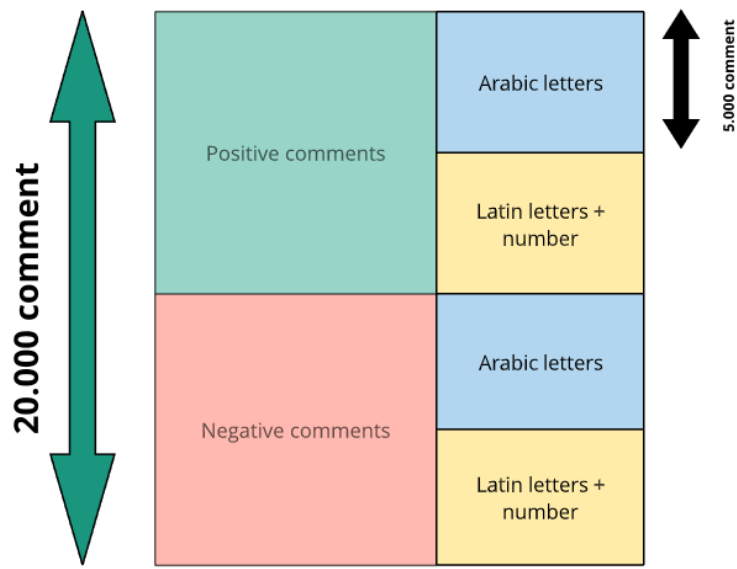

Fig. 2. the contents MYC Corpus.

Table 3. Examples of Moroccan dialect texts in Arabic letters and Latin letters

| Text in English | Using Arabic letters | Using Latin letters |
|---|---|---|
| this is beautiful | هادشي زوين | Hadchi zuin |
| I really like this product | عجبني هاد البرودوي | 3jbni had lproduit |

As a result, we managed to create the largest publicly available dataset for MD and the only one that covers both scripting styles of the language. Table 4 show the comparison of various Moroccan datasets and their features.

Table 4. Comparison of available Moroccan datasets.

| Reference | Dataset | Size | Scripting types | Availability |
|---|---|---|---|---|
| [11,2020] | Moroccan Sentiment Analysis Corpus (MSAC) | 2.000 | Arabic script | Public |
| [16,2020] | Moroccan Sentiment Twitter Dataset | 12.000 | Arabic script | Private |
| [17,2021] | Moroccan Arabic Corpus (MAC) | 8.360 | Arabic script | Public |
| [18,2022] | Modeling, Simulation and Data Analysis Dataset | 4.855 | Arabic script | Private |
| **This work** | **Moroccan YouTube Corpus (MYC)** | **20.000** | **Arabic script and Latin Script** | **Public** |

## 5.Adopted Approaches for Machine Learning models

We propose a comparison of a variety of models, ranging from classical and standard models to newly proposed new models. We looked into DL models, such as CNN and LSTM, in addition to so-called classic models like SVM, NB, and KNN.

in this section we explain the steps performed to create a model for MD sentiment analyses. we reached that by applying various text preprocessing techniques on the YouTube comments and then training some ML classifiers. The whole process composes of two steps. First the training step where our model classifiers learn from pre labeled comments. And second the testing step, where unlabeled comments will be classified by the trained classifiers and compare the results and accuracy of each classifier and preprocessing techniques.
    The whole methodology main stages are: preprocessing steps, generating features, comparing several supervised machine-learning classification methods, and presenting their results.

## 5.1 Preprocessing data:

Numerous examinations and studies uncovered and talked about that pre-processing of textual content can enhance the performance of textual content classification [23][24], hence we considered various pre-processing techniques including:

1) Data cleaning: Data cleaning is removing all noisy data in comments such as publicity of a YouTube channels, links, special characters, user-names, dates, video time tags (Example: 2:32), URLs and numbers but we only deleted numbers written separately since some Latin scripted words use numbers.

2) Text Tokenization: Tokenization is a task that consisted in separating each word from another, because it is easy to analyze each word by itself.

3) Normalization: Removing diacritics, normalizing the shapes of identical Arabic letters, and omitting duplicated characters are the steps of normalization that we used for the following reasons before conducting our experiments.

- Diacritics are used to alter the pronunciation and often the meaning of the words in the MSA language and written above the letters of the word. However, diacritics are rarely used in the dialects used on social media platforms, and they are only used for decorative purposes in most cases.
- Similar Arabic Letters Normalization: Alternatively, we typically prefer to write similar Arabic letters, especially the Hamza (ء) when it exists with the letter of Aleph (ا), which is pronounced differently depending on its location. Generally, because we know that the text context indicates its intended meaning, we are not worried about its correct spelling. (e.g., we write الى instead of إلى.)
- In addition, the Ta marbuta (ة) which appears only at the end of words and has a sort of similar shape to the letter Ha (ه) also at the end of the word, web users sometimes interchangeably write the Ha and the Ta marbuta. Therefore, we converted all the shapes of Aleph with the Hamza or the Madda to a normal Aleph, and all Ta marbuta to Ha letter.
- Duplicate Letters Removal: The users of social websites often write the same letters many times consecutively to reinforce the significance of the words, or by an error (e.g., tooop or جمييل). To solve this dilemma, a simple solution is to delete all of the related redundant letters. It is also important to compact punctuations (such as commas and dots) and repeated spaces and replace them with one spatial letter.

4) Emojis treatment: Most previous research when it comes to sentiment analyses, they consider Emojis as noisy data and just delete them from the treated texts. In our work we decided to proceed with two different datasets one that contains Emojis and one where we deleted all emojis and then test with both datasets.

5) Delete stop words: Stop words are words that are filtered before data processing because they contain no additional meaning to the text (e.g. a/an/the…). There are many tools and libraries to delete stop words from many languages such English, French and even MSA but none of them had a list of MD stop words. Therefore, we created manually two list of MD stop words, one written with Arabic letters and the other one with Latin letters.

6) Filtering Repeated Comments: All comments were processed after implementing all these standardization measures by eliminating duplicate comments to ensure the uniqueness of the dataset in the next stages.

## 5.2 Feature extraction:

Feature extraction represents an essential process for optimizing ML models. This avoids the problem of overfitting which reduces the dimensionality of the data. Therefore, selection is made among the most relevant features, which significantly improves accuracy and reduces the time required in training the model.

We used the well-known TF-IDF algorithm [25], which not only calculates word occurrence but also assigns an individual TF-IDF score to each word according to the ratio of the two measurements: TF represents the frequency of a word in a document, and IDF calculates its importance in the dataset, where the words that appear least frequently in the document are the most important words for classification purposes.

To use DL models, the approach requires the creation of word embeddings. Word2vec is a famous word embedding model developed at Google in 2013. It integrates a neural network layer to predict words adjacent to the target word in the case of the skip gram architecture, or the word from its neighbors in the case of the CBOW (Continuous bag of words). The input and output of Word2vec is an encoding of the vocabulary in the data set. During training, a window moves through the corpus [26], training the neural network to predict surrounding or target words for each word by assigning probabilities to those words. After training, the network layer is the vector representation of a word. In our case, we trained our own word embeddings using the Word2vec algorithm on our dataset.

## 6. Experimental results and discussions

### 6.1 Comparative study of created models

In this section we present several experiments for SA of Moroccan texts. To determine the most precise ML algorithms for SA tasks. For CNN and LSTM classifiers we decided to set the hyperparameters of the network architecture used by Oswal [27] and Nagy [28]. The evaluation metrics used in this experiment are accuracy, precision, recall and F-measure.

Table 5. Results without pre-processing.

| Feature | SVM | NB | KNN | CNN | LSTM |
|---|---|---|---|---|---|
| Accuracy (%) | 73.5 | 71.5 | 68.6 | 85 | 84 |
| Precision (%) | 72.9 | 71 | 68 | 84 | 83.7 |
| Recall (%) | 73 | 71 | 69 | 84 | 84 |
| F-measure (%) | 72.8 | 70.8 | 69 | 85 | 83.5 |

Table 5 presents the results achieved from different classifiers trained on raw data in terms of accuracy, recall and F-measure.

We can compare the results of the tested ML classifiers trained on raw data using the experimental results shown in Table 5. In almost all of the evaluation measures, the results show that SVM outperformed the NB classifier. It was accurate to 73.5%. KNN performed worse than NB during the experiment.
Experiments comparing DL and single models show that DL improves classification efficiency in terms of accuracy and other measures. Indeed, the CNN and LSTM algorithms performed admirably and outperformed the classic models (SVM, NB and KNN). The CNN and LSTM algorithms ensure the highest accuracy with 85% and 84%, respectively.

When comparing the results obtained with the results of ML and DL without pre-processing, the best performance of all classifiers is achieved when the pre-processing stage is applied. In fact, as can be seen from Tables 6 and 7, removing stop words and emojis gives the best results in the classifiers.
These results confirm that the pre-processing step is necessary in order to improve the accuracy and performance of the MD, since pre-processing is a good way to reduce the noise in the text. We also believe that creating lists of Moroccan stop words helped further to increase the accuracy of our model rather than using MSA stop words alone.

Table 6. results with pre-processing and without deleting emojis.

| Feature | SVM | NB | KNN | CNN | LSTM |
| --- | --- | --- | --- | --- | --- |
| Accuracy (%) | 76.8 | 76 | 64 | 87 | 85.5 |
| Precision (%) | 76.7 | 75 | 64 | 86 | 85 |
| Recall (%) | 75.1 | 75 | 63.2 | 86.8 | 85.2 |
| F-measure (%) | 76 | 75 | 63.7 | 86.7 | 85.2 |

Table 7. results with pre-processing and with deleted emojis.

| Feature | SVM | NB | KNN | CNN | LSTM |
| --- | --- | --- | --- | --- | --- |
| Accuracy (%) | 84.7 | 83.3 | 75 | **92.4** | 87.2 |
| Precision (%) | 84 | 83 | 75 | 89 | 87 |
| Recall (%) | 84.5 | 83.2 | 74 | 89.6 | 87.2 |
| F-measure (%) | 84.2 | 83 | 75.2 | 92 | 89 |

We would like to note that we only used unigram representation in all datasets since the bigram showed poor results compared to unigrams. The key behind the good results of unigram approach, which allows for greater accuracy, is that in the MD, negation is not done by adding a negation word such as "not" in English or "ne pas" in French, but by adding a suffix or prefix to the word, for example the word "زوين" (beautiful) its negation is " مزوينش" (not beautiful).

Additionally, we observed that the impact of deleting emojis was due to the biased of our dataset when it comes to emojis, after investigating we managed to notice that the ratio of positive emojis in our dataset exceed by a great amount the ratio of negative emojis.

We conclude that, in comparison to other methods, DL models are recommended for Arabic and Arabic dialects sentiment classification because they ensure high accuracy and performance. However, this solution has a drawback in that it takes longer than a single model to complete the training phase.

**6.2 Testing the CNN pre- trained model on other Moroccan datasets**

In this research study, we conducted extensive testing of our trained SA classifier on a diverse set of publicly available datasets of MD beyond the one used for training, although both of the datasets used for testing contain only one scripting type of MD. This approach serves as a critical evaluation step to assess the generalizability and robustness of our model and dataset. By subjecting our model to a variety of data sources, we aim to gauge its performance in real-world scenarios and ensure that it can effectively handle data variations and linguistic nuances. This comprehensive testing not only validates the model's adaptability but also contributes to the broader understanding of its applicability in practical, cross-domain SA tasks, strengthening the credibility and relevance of our research findings.

To perform this comparative study, we used the two public datasets previously mentioned MAC and MSAC, by applying our best model from the previous experiment on 20% of each dataset as testing sets to properly compare to the others researches results, we focused on the accuracy as a metric for this study. The results of our experiment are more detailed in table 8.

Table 8. comparative results of testing our model using other datasets.

|            | Datasets |     |
| ---------- | -------- | --- |
| Models     | MSAC     | MAC |
| Our Model  | **99**   | **95** |
| MSAC Model | 99       | --- |
| MAC Model  | ---      | 93  |

In the comparative analysis table 8, it is evident that our trained model exhibits a noteworthy advantage over MAC by outperforming it by a substantial margin of 2% accuracy. Furthermore, our model achieves an accuracy score equivalent to that of MSAC, we believe the small size of their dataset contributes to how high the accuracy is. This parity with MSAC underscores the competitive strength of our dataset in delivering consistent and robust SA results, showcasing its proficiency in sentiment classification tasks across different Moroccan datasets.

## Conclusion

In conclusion, this research paper represents a significant contribution to the field of SA, specifically focusing on the MD. We have successfully achieved the following key milestones:

Firstly, we created a comprehensive SA dataset, comprising 20,000 MD texts written in both Arabic and Latin letters. In addition, we managed to create lists of stop words adapted to the MD. This dataset and resource are a valuable asset to the research community, addressing the scarcity of publicly available resources for SA in the MD.

Secondly, we designed and implemented SA models, integrating both traditional ML and DL techniques. Remarkably, our CNN model achieved an impressive accuracy rate of 92%. This result underlines the robustness and effectiveness of our model in accurately capturing sentiment in MD texts.

In addition, to validate the credibility and generalizability of our model and dataset, we carried out a thorough evaluation on other publicly available MD datasets. The results of this comparative analysis demonstrated the superiority of our approach, attesting to the quality and reliability of our dataset and the competence of our SA model.

In summary, this research strives to bridge the existing gap in the field of natural language processing of dialectal texts when it comes to different types of scripts, by offering a comprehensive SA dataset and a highly accurate SA model adapted to the MD. Our results not only constitute an essential resource for researchers in this field, but also demonstrate the effectiveness of our methodology, thus advancing SA for the MD. This work lays a solid foundation for future research and applications in the field of SA for the Arabic dialect.

**DECLARATION OF COMPETING INTEREST:**

The authors declare that they have no known competing financial interests or personal relationships that could have appeared to influence the work reported in this paper.

**FUNDING:**

This research did not receive any specific grant from funding agencies in the public, commercial, or not-for-profit sectors.